\documentclass[10pt,twocolumn,letterpaper]{article}

\usepackage{cvpr}
\usepackage{times}
\usepackage{epsfig}
\usepackage{graphicx}
\usepackage{amsmath}
\usepackage{amssymb}
\usepackage{multirow}
\usepackage{subcaption}
\usepackage{verbatim}
\usepackage{placeins}

\newcommand{\bR}{\mathbf{R}}

\newcommand{\bt}{\mathbf{t}}

\newcommand{\bT}{\mathbf{T}}

\newcommand{\bH}{\mathbf{H}}

\newcommand{\be}{\mathbf{e}}

\def\be {\begin{equation}}
\def\ee {\end{equation}}
\def\beas {\begin{eqnarray*}}
\def\eeas {\end{eqnarray*}}
\def\bea {\begin{eqnarray}}
\def\eea {\end{eqnarray}}

\usepackage{enumitem}
\usepackage[pagebackref=true,breaklinks=true,letterpaper=true,colorlinks,bookmarks=false]{hyperref}

\cvprfinalcopy % *** Uncomment this line for the final submission

\def\cvprPaperID{704} % *** Enter the CVPR Paper ID here

\ifcvprfinal\fi
\begin{document}

%%%%%%%%% TITLE
\title{Multi-View 3D Object Detection Network for Autonomous Driving}

\author{Xiaozhi Chen$^1$, Huimin Ma$^1$, Ji Wan$^2$, Bo Li$^2$, Tian Xia$^2$\\
$^1$Department of Electronic Engineering, Tsinghua University\\
$^2$Baidu Inc.\\
{\tt\small \{chenxz12@mails., mhmpub@\}tsinghua.edu.cn, }
{\tt\small \{wanji, libo24, xiatian\}@baidu.com}
%\and
}

\maketitle
\thispagestyle{empty}

%%%%%%%%% ABSTRACT
\begin{abstract}
This paper aims at high-accuracy 3D object detection in autonomous driving scenario.
We propose Multi-View 3D networks (MV3D), a sensory-fusion framework that takes both LIDAR point cloud and RGB images as input and predicts oriented 3D bounding boxes.
We encode the sparse 3D point cloud with a compact multi-view representation.
The network is composed of two subnetworks: one for 3D object proposal generation and another for multi-view feature fusion.
The proposal network generates 3D candidate boxes efficiently from the bird's eye view representation of 3D point cloud.
We design a deep fusion scheme to combine region-wise features from multiple views and enable interactions between intermediate layers of different paths.
Experiments on the challenging KITTI benchmark show that our approach outperforms the state-of-the-art by around 25\% and 30\% AP on the tasks of 3D localization and 3D detection.
In addition, for 2D detection, our approach obtains 10.3\% higher AP than the state-of-the-art on the hard data among the LIDAR-based methods.

\end{abstract}

%%%%%%%%% BODY TEXT
\section{Introduction}

3D object detection plays an important role in the visual perception system of Autonomous driving cars.
Modern self-driving cars are commonly equipped with multiple sensors, such as LIDAR and cameras.
Laser scanners have the advantage of accurate depth information while cameras preserve much more detailed semantic information.
The fusion of LIDAR point cloud and RGB images should be able to achieve higher performance and safty to self-driving cars.

The focus of this paper is on 3D object detection utilizing both LIDAR and image data.
We aim at highly accurate 3D localization and recognition of objects in the road scene.
Recent LIDAR-based methods place 3D windows in 3D voxel grids to score the point cloud~\cite{vote3d, vote3deep} or apply convolutional networks to the front view point map in a dense box prediction scheme~\cite{velofcn}.
Image-based methods~\cite{XiaozhiNIPS15, XiaozhiCVPR16} typically first generate 3D box proposals and then perform region-based recognition using the Fast R-CNN~\cite{fastrcnn} pipeline.
Methods based on LIDAR point cloud usually achieve more accurate 3D locations while image-based methods have higher accuracy in terms of 2D box evaluation.
~\cite{AGonzalez2016, multilevelTIP11} combine LIDAR and images for 2D detection by employing early or late fusion schemes.
However, for the task of 3D object detection, which is more challenging, a well-designed model is required to make use of the strength of multiple modalities.

\begin{figure*}[t!]
	\begin{center}
        \vspace{-1mm}
		\includegraphics[width=0.95\linewidth,trim = 0mm 0mm 0mm 0mm, clip]{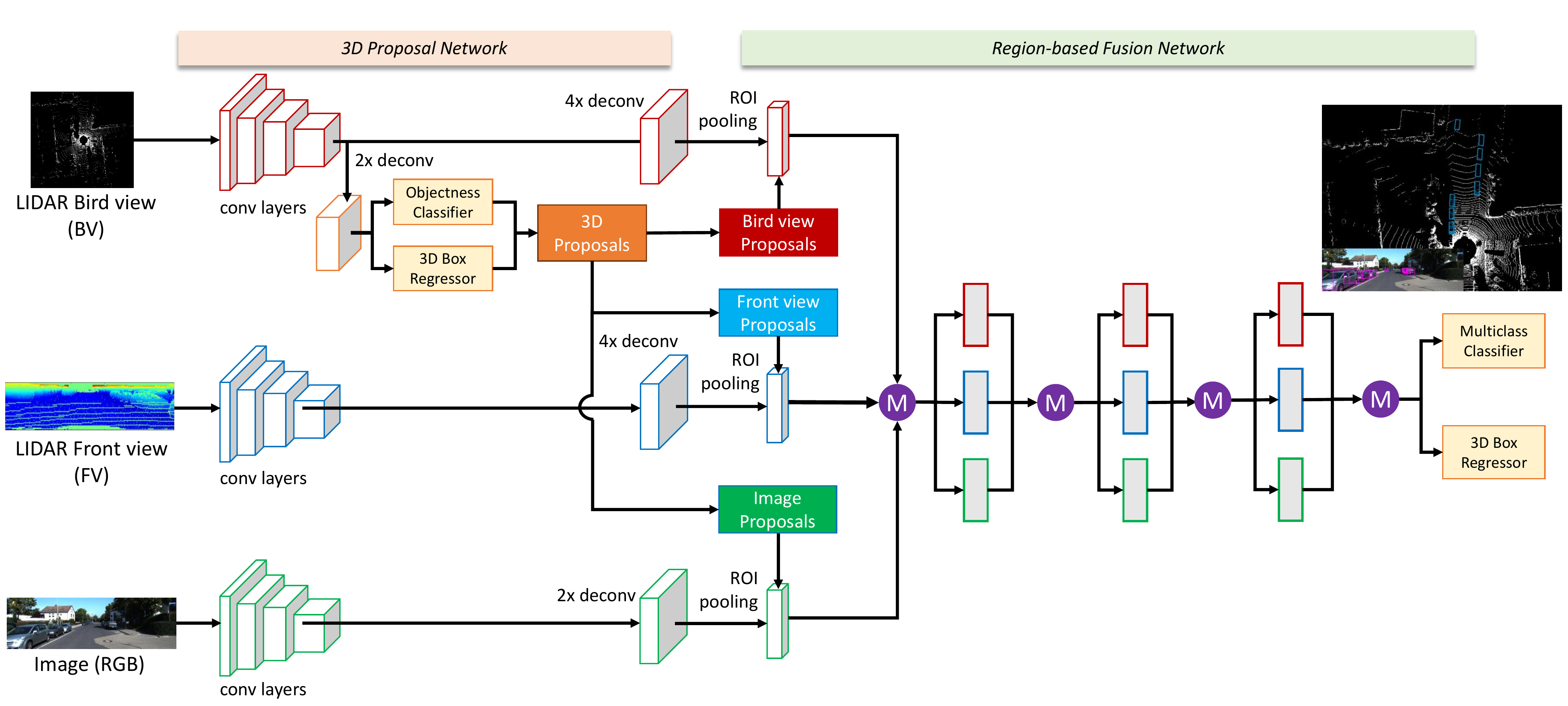}
        \vspace{-2.5mm}
        \caption{{\bf Multi-View 3D object detection network (MV3D):}
        The network takes the bird's eye view and front view of LIDAR point cloud as well as an image as input.
        It first generates 3D object proposals from bird's eye view map and project them to three views.
        A deep fusion network is used to combine region-wise features obtained via ROI pooling for each view.
        The fused features are used to jointly predict object class and do oriented 3D box regression.}
		\label{fig:net}
	\end{center}
    \vspace{-5mm}
\end{figure*}

In this paper, we propose a Multi-View 3D object detection network ({\bf MV3D}) which takes multimodal data as input and predicts the full 3D extent of objects in 3D space.
The main idea for utilizing multimodal information is to perform region-based feature fusion.
We first propose a multi-view encoding scheme to obtain a compact and effective representation for sparse 3D point cloud.
As illustrated in Fig.~\ref{fig:net}, the multi-view 3D detection network consists of two parts: a \emph{3D Proposal Network} and a \emph{Region-based Fusion Network}.
The 3D proposal network utilizes a bird's eye view representation of point cloud to generate highly accurate 3D candidate boxes.
The benefit of 3D object proposals is that it can be projected to any views in 3D space.
The multi-view fusion network extracts region-wise features by projecting 3D proposals to the feature maps from mulitple views.
We design a deep fusion approach to enable interactions of intermediate layers from different views.
Combined with drop-path training~\cite{fractalnet} and auxiliary loss, our approach shows superior performance over the early/late fusion scheme.
Given the multi-view feature representation, the network performs \emph{oriented} 3D box regression which predict accurate 3D location, size and orientation of objects in 3D space.

We evaluate our approach for the tasks of 3D proposal generation, 3D localization, 3D detection and 2D detection on the challenging KITTI~\cite{kitti} object detection benchmark.
Experiments show that our 3D proposals significantly outperforms recent 3D proposal methods 3DOP~\cite{XiaozhiNIPS15} and Mono3D~\cite{XiaozhiCVPR16}.
In particular, with only 300 proposals, we obtain 99.1\% and 91\% \emph{3D recall} at Intersection-over-Union (IoU) threshold of 0.25 and 0.5, respectively.
The LIDAR-based variant of our approach achieves around 25\% higher accuracy in 3D localization task and 30\% higher 3D Average Precision (AP) in the task of 3D object detection.
It also outperforms all other LIDAR-based methods by 10.3\% AP for 2D detection on KITTI's hard test set.
When combined with images, further improvements are achieved over the LIDAR-based results.

\section{Related Work}
We briefly review existing work on 3D object detection from point cloud and images, multimodal fusion methods and 3D object proposals.

\vspace{-0.3cm}
\paragraph{3D Object Detection in Point Cloud.}
Most existing methods encode 3D point cloud with voxel grid representation.
Sliding Shapes~\cite{slidingshape} and Vote3D~\cite{vote3d} apply SVM classifers on 3D grids encoded with geometry features.
Some recently proposed methods~\cite{deepslidingshape,vote3deep,li3dfcn} improve feature representation with 3D convolutions.networks, which, however require expensive computations.
In addition to the 3D voxel representation, VeloFCN~\cite{velofcn} projects point cloud to the front view, obtaining a 2D point map.
They apply a fully convolutional network on the 2D point map and predict 3D boxes densely from the convolutional feature maps.
~\cite{su2015multi, qi2016volumetric, fusionnet} investigate volumetric and multi-view representation of point cloud for 3D object classification.
In this work, we encode 3D point cloud with multi-view feature maps, enabling region-based representation for multimodal fusion.

\begin{figure*}[t!]
	\begin{center}
		\includegraphics[width=0.95\linewidth,trim = 0mm 0mm 0mm 0mm, clip]{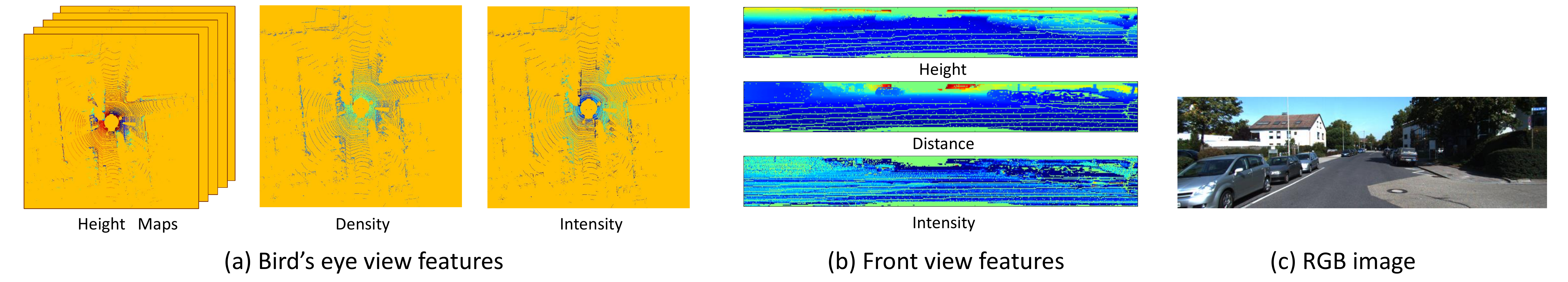}
		\vspace{-2.5mm}
		\caption{Input features of the MV3D network.}
		\label{fig:feat}
	\end{center}
    \vspace{-5mm}
\end{figure*}

\vspace{-0.4cm}
\paragraph{3D Object Detection in Images.}
3DVP~\cite{xiangcvpr15} introduces 3D voxel patterns and employ a set of ACF detectors to do 2D detection and 3D pose estimation.
3DOP~\cite{XiaozhiNIPS15} reconstructs depth from stereo images and uses an energy minimization approach to generate 3D box proposals, which are fed to an R-CNN~\cite{fastrcnn} pipeline for object recognition.
While Mono3D~\cite{XiaozhiCVPR16} shares the same pipeline with 3DOP, it generates 3D proposals from monocular images.
~\cite{ZiaPAMI13, ZiaCVPR14} introduces a detailed geometry representation of objects using 3D wireframe models.
To incorporate temporal information, some work\cite{Dhiman2016A, Song2015Joint} combine structure from motion and ground estimation to lift 2D detection boxes to 3D bounding boxes.
Image-based methods usually rely on accurate depth estimation or landmark detection.
Our work shows how to incorporate LIDAR point cloud to improve 3D localization.

\vspace{-0.4cm}
\paragraph{Multimodal Fusion}
Only a few work exist that exploit multiple modalities of data in the context of autonomous driving.
~\cite{AGonzalez2016} combines images, depth and optical flow using a mixture-of-experts framework for 2D pedestrian detection.
~\cite{multilevelTIP11} fuses RGB and depth images in the early stage and trains pose-based classifiers for 2D detection.
In this paper, we design a deep fusion approach inspired by FractalNet~\cite{fractalnet} and Deeply-Fused Net~\cite{dfn}.
In FractalNet, a base module is iteratively repeated to construct a network with exponentially increasing paths.
Similarly,~\cite{dfn} constructs deeply-fused networks by combining shallow and deep subnetworks.
Our network differs from them by using the same base network for each column and adding auxiliary paths and losses for regularization.

\vspace{-0.4cm}
\paragraph{3D Object Proposals}
Similarly to 2D object proposals~\cite{van2011segmentation, zitnick2014edge, CarreiraCpmcPAMI2012}, 3D object proposal methods generate a small set of 3D candidate boxes in order to cover most of the objects in 3D space.
To this end, 3DOP~\cite{XiaozhiNIPS15} designs some depth features in stereo point cloud to score a large set of 3D candidate boxes.
Mono3D~\cite{XiaozhiCVPR16} exploits the ground plane prior and utilizes some segmentation features to generate 3D proposals from a single image.
Both 3DOP and Mono3D use hand-crated features.
Deep Sliding Shapes~\cite{deepslidingshape} exploits more powerful deep learning features.
However, it operates on 3D voxel grids and uses computationally expensive 3D convolutions.
We propose a more efficient approach by introducing a bird's eye view representation of point cloud and employing 2D convolutions to generate accurate 3D proposals.

\section{MV3D Network}

The MV3D network takes a multi-view representation of 3D point cloud and an image as input.
It first generates 3D object proposals from the bird's eye view map and deeply fuses multi-view features via region-based representation.
The fused features are used for category classification and oriented 3D box regression.

\vspace{-0.0cm}
\subsection{3D Point Cloud Representation}
Existing work usually encodes 3D LIDAR point cloud into a 3D grid~\cite{vote3d, vote3deep} or a front view map~\cite{velofcn}.
While the 3D grid representation preserves most of the raw information of the point cloud, it usually requires much more complex computation for subsequent feature extraction.
We propose a more compact representation by projecting 3D point cloud to the bird's eye view and the front view.
Fig.~\ref{fig:feat} visualizes the point cloud representation.

\vspace{-0.4cm}
\paragraph{Bird's Eye View Representation.}
The bird's eye view representation is encoded by height, intensity and density.
We discretize the projected point cloud into a 2D grid with resolution of 0.1m.
For each cell, the height feature is computed as the maximum height of the points in the cell.
To encode more detailed height information, the point cloud is devided equally into $M$ slices.
A height map is computed for each slice, thus we obtain $M$ height maps.
The intensity feature is the reflectance value of the point which has the maximum height in each cell.
The point cloud density indicates the number of points in each cell.
To normalize the feature, it is computed as $min(1.0, \frac{log(N+1)}{log(64)})$, where $N$ is the number of points in the cell.
Note that the intensity and density features are computed for the whole point cloud while the height feature is computed for $M$ slices, thus in total the bird's eye view map is encoded as $(M+2)$-channel features.

\vspace{-0.4cm}
\paragraph{Front View Representation.}
Front view representation provides complementary information to the bird's eye view representation.
As LIDAR point cloud is very sparse, projecting it into the image plane results in a sparse 2D point map.
Instead, we project it to a cylinder plane to generate a dense front view map as in~\cite{velofcn}.
Given a 3D point $p=(x, y, z)$, its coordinates $p_{fv}=(r, c)$ in the front view map can be computed using

\begin{equation}
\begin{aligned}
c & =\lfloor \text{atan2}(y, x)/\Delta\theta] \rfloor \\
r & =\lfloor \text{atan2}(z, \sqrt{x^2+y^2})/\Delta\phi \rfloor,
\end{aligned}
\label{eq:fv}
\end{equation}
where $\Delta\theta$ and $\Delta\phi$ are the horizontal and vertical resolution of laser beams, respectively.
We encode the front view map with three-channel features, which are height, distance and intensity, as visualized in Fig.~\ref{fig:feat}.

\subsection{3D Proposal Network}
Inspired by Region Proposal Network (RPN) which has become the key component of the state-of-the-art 2D object detectors~\cite{fasterrcnn}, we first design a network to generate 3D object proposals.
We use the bird's eye view map as input.
In 3D object detection, The bird's eye view map has several advantages over the front view/image plane.
First, objects preserve physical sizes when projected to the bird's eye view, thus having small size variance, which is not the case in the front view/image plane.
Second, objects in the bird's eye view occupy different space, thus avoiding the occlusion problem.
Third, in the road scene, since objects typically lie on the ground plane and have small variance in vertical location, the bird's eye view location is more crucial to obtaining accurate 3D bounding boxes.
Therefore, using explicit bird's eye view map as input makes the 3D location prediction more feasible.

\iffalse
\begin{figure*}[t!]
	\begin{center}
		\vspace{-1mm}
		\includegraphics[width=0.9\linewidth,trim = 0mm 0mm 0mm 0mm, clip]{figs/fusion2.pdf}
		\vspace{-2.5mm}
		\caption{{\bf Architectures of different fusion schemes:} We instantiate the join nodes in early/late fusion with \emph{concatenation} operation, and deep fusion with \emph{element-wise mean} operation.}
		\label{fig:fusion}
	\end{center}
    \vspace{-5mm}
\end{figure*}
\fi

\begin{figure}[t!]
	\begin{center}
		\vspace{-1mm}
		\includegraphics[width=0.9\linewidth,trim = 0mm 0mm 0mm 0mm, clip]{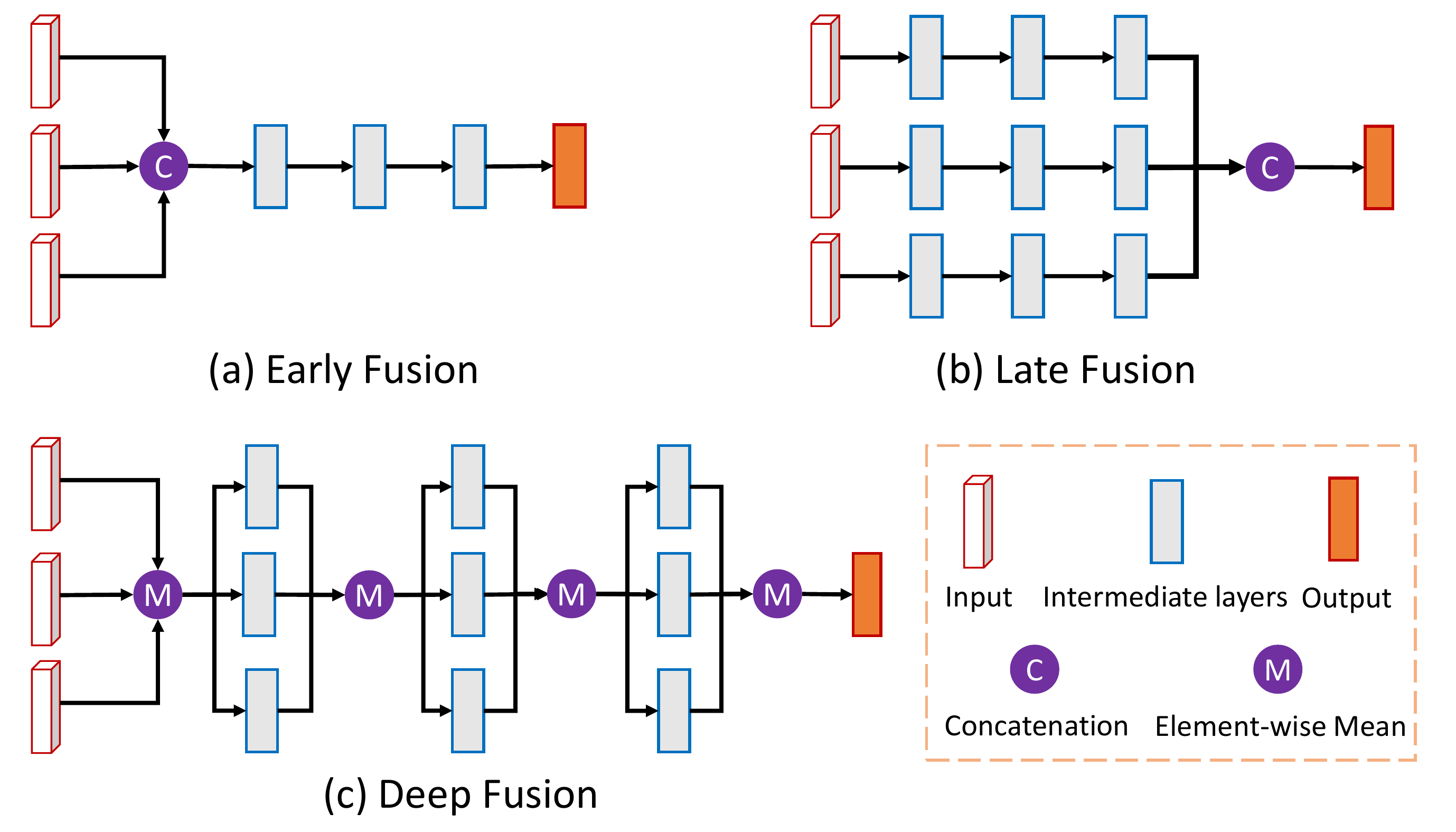}
		\vspace{-2.5mm}
		\caption{{\bf Architectures of different fusion schemes:} We instantiate the join nodes in early/late fusion with \emph{concatenation} operation, and deep fusion with \emph{element-wise mean} operation.}
		\label{fig:fusion}
	\end{center}
    \vspace{-5mm}
\end{figure}

Given a bird's eye view map. the network generates 3D box proposals from a set of 3D prior boxes.
Each 3D box is parameterized by $(x, y, z, l, w, h)$, which are the center and size (in meters) of the 3D box in LIDAR coordinate system.
For each 3D prior box, the corresponding bird's eye view anchor $(x_{bv}, y_{bv}, l_{bv}, w_{bv})$ can be obtained by discretizing $(x, y, l, w)$.
We design $N$ 3D prior boxes by clustering ground truth object sizes in the training set.
In the case of car detection, $(l, w)$ of prior boxes takes values in $\{(3.9, 1.6), (1.0, 0.6)\}$, and the height $h$ is set to 1.56m.
By rotating the bird's eye view anchors 90 degrees, we obtain $N=4$ prior boxes.
$(x, y)$ is the varying positions in the bird's eye view feature map, and $z$ can be computed based on the camera height and object height.
We do not do orientation regression in proposal generation, whereas we left it to the next prediction stage.
The orientations of 3D boxes are restricted to $\{0^{\circ}, 90^{\circ}\}$, which are close to the actual orientations of most road scene objects.
This simplification makes training of proposal regression easier.

With a disretization resolution of 0.1m, object boxes in the bird's eye view only occupy 5$\sim$40 pixels.
Detecting such extra-small objects is still a difficult problem for deep networks.
One possible solution is to use higher resolution of the input, which, however, will require much more computation.
We opt for feature map upsampling as in~\cite{mscnn}.
We use 2x bilinear upsampling after the last convolution layer in the proposal network.
In our implementation, the front-end convolutions only proceed three pooling operations, i.e., 8x downsampling.
Therefore, combined with the 2x deconvolution, the feature map fed to the proposal network is 4x downsampled with respect to the bird's eye view input.

We do 3D box regression by regressing to $\bt=(\Delta x, \Delta y, \Delta z, \Delta l, \Delta w, \Delta h)$, similarly to RPN~\cite{fasterrcnn}.
$(\Delta x, \Delta y, \Delta z)$ are the center offsets normalized by anchor sizes,
and $(\Delta l, \Delta w, \Delta h)$ are computed as $\Delta s = log\frac{s_{\text{GT}}}{s_{\text{anchor}}}, s \in \{l, w, h\}$.
we use a multi-task loss to simultaneously classify object/background and do 3D box regression.
In particular, we use class-entropy for the ``objectness" loss and Smooth $\ell_1$~\cite{fastrcnn} for the 3D box regression loss.
Background anchors are ignored when computing the box regression loss.
During training, we compute the IoU overlap between anchors and ground truth bird's eye view boxes.
An anchor is considered to be positive if its overlap is above 0.7, and negative if the overlap is below 0.5.
Anchors with overlap in between are ignored.

Since LIDAR point cloud is sparse, which results in many empty anchors, we remove all the empty anchors during both training and testing to reduce computation.
This can be achieved by computing an integral image over the point occupancy map.

For each non-empty anchor at each position of the last convolution feature map, the network generates a 3D box.
To reduce redundancy, we apply Non-Maximum Suppression (NMS) on the bird's eye view boxes.
Different from~\cite{deepslidingshape}, we did not use 3D NMS because objects should occupy different space on the ground plane.
We use IoU threshold of 0.7 for NMS. The top 2000 boxes are kept during training, while in testing, we only use 300 boxes.

\begin{figure}[t!]
	\begin{center}
		\vspace{-1mm}
		\includegraphics[width=0.98\linewidth,trim = 0mm 0mm 0mm 0mm, clip]{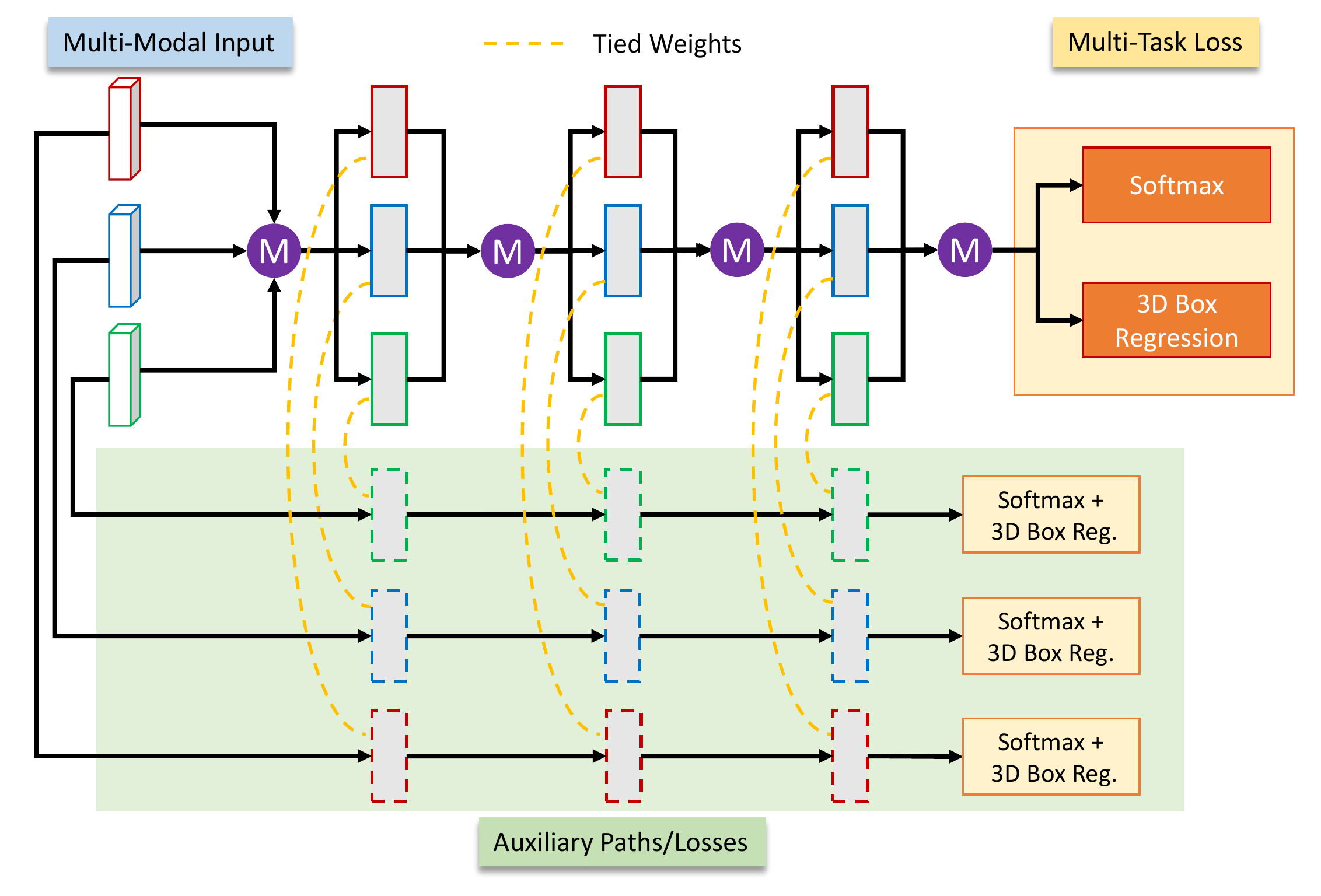}
		\vspace{-2.5mm}
		\caption{{\bf Training strategy for the Region-based Fusion Network:}
        During training, the bottom three paths and losses are added to regularize the network. The auxiliary layers share weights with the corresponding layers in the main network.}
		\label{fig:fusion_net}
		\vspace{-6mm}
	\end{center}
\end{figure}

\begin{figure*}[t!]
\begin{center}
\vspace{-1mm}
\begin{tabular}{ccc}
\raisebox{-0.38\height}{\includegraphics[width=0.27\linewidth,trim = 5mm 0mm 0mm 0mm, clip]{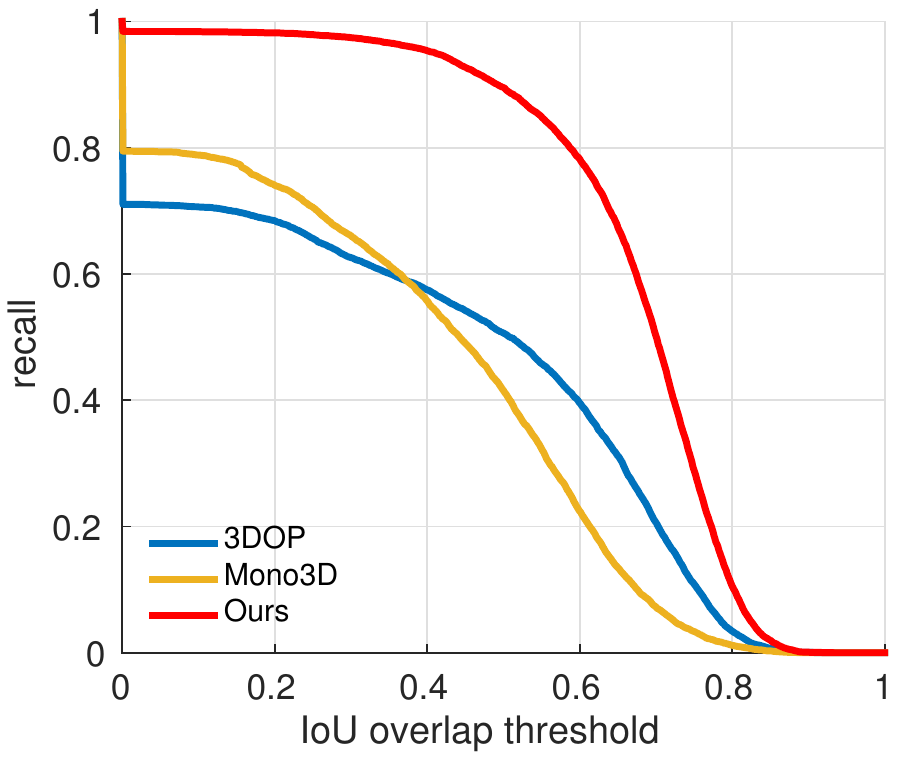}}&
\raisebox{-0.38\height}{\includegraphics[width=0.27\linewidth,trim = 5mm 0mm 0mm 0mm, clip]{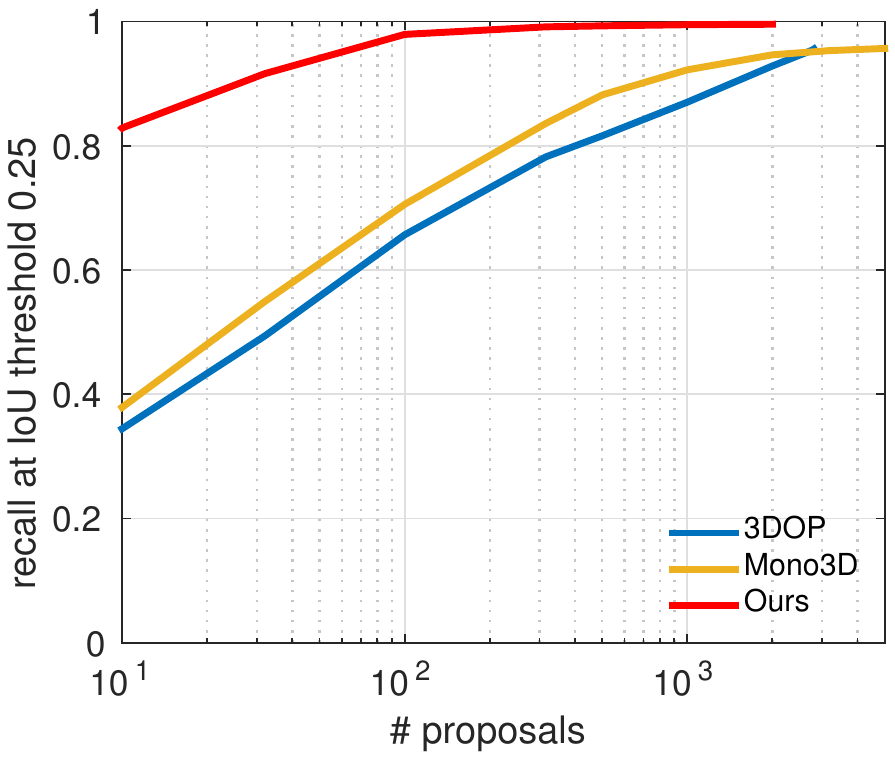}}&
\raisebox{-0.38\height}{\includegraphics[width=0.27\linewidth,trim = 5mm 0mm 0mm 0mm, clip]{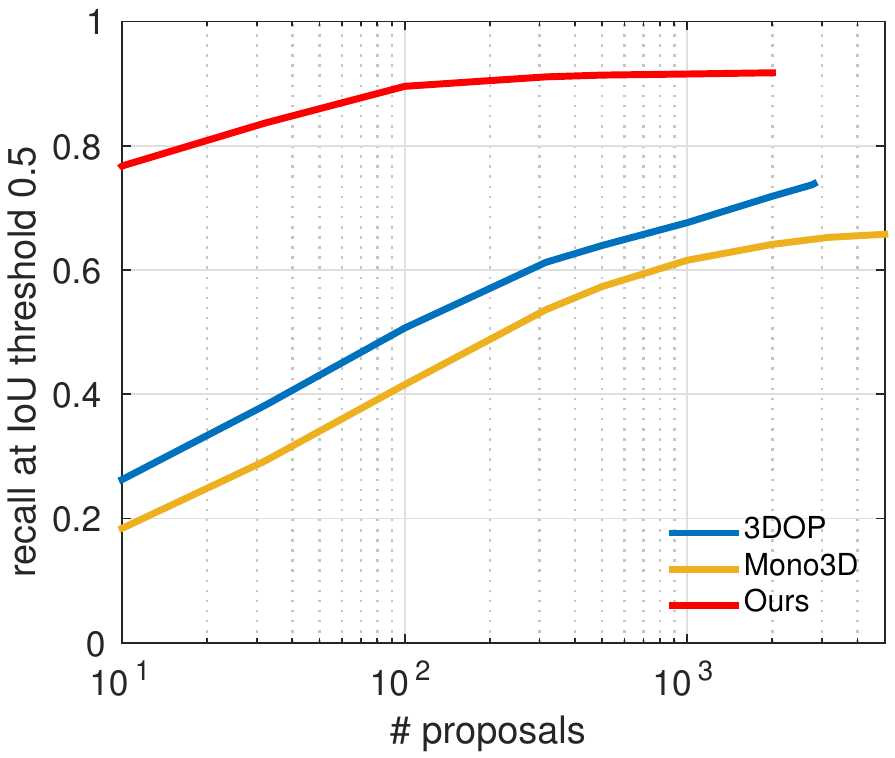}}\\
\end{tabular}
\vspace{-2.5mm}
\caption{\textbf{3D bounding box Recall:} From left to right: Recall vs IoU using 300 proposals, Recall vs \#Proposals at IoU threshold of 0.25 and 0.5 respectively. Recall are evaluated on moderate data of KITTI \emph{validation} set.}
\label{fig:3d-recall}
\vspace{-5mm}
\end{center}
\end{figure*}

\subsection{Region-based Fusion Network}
We design a region-based fusion network to effectively combine features from multiple views and jointly classify object proposals and do oriented 3D box regression.

\vspace{-0.4cm}
\paragraph{Multi-View ROI Pooling.}
Since features from different views/modalities usually have different resolutions, we employ ROI pooling~\cite{fastrcnn} for each view to obtain feature vectors of the same length.
Given the generated 3D proposals, we can project them to any views in the 3D space.
In our case, we project them to three views, i.e., bird's eye view (BV), front view (FV), and the image plane (RGB).
Given a 3D proposal $p_{\text{3D}}$, we obtain ROIs on each view via:
\begin{equation}
\text{ROI}_v = \bT_{\text{3D} \to v}(p_{\text{3D}}), v \in \{\text{BV}, \text{FV}, \text{RGB}\}
\end{equation}
where $\bT_{\text{3D} \to v}$ denotes the tranformation functions from the LIDAR coordinate system to the bird's eye view, front view, and the image plane, respectively.
Given an input feature map $x$ from the front-end network of each view, we obtain fixed-length features $f_v$ via ROI pooling:
\begin{equation}
f_v = \bR(x, \text{ROI}_v), v \in \{\text{BV}, \text{FV}, \text{RGB}\}.
\end{equation}

\vspace{-0.4cm}
\paragraph{Deep Fusion.}
To combine information from different features, prior work usually use \emph{early fusion}~\cite{mscnn} or \emph{late fusion}~\cite{deepslidingshape, hallCVPR16}.
Inspired by~\cite{fractalnet, dfn}, we employ a \emph{deep fusion} approach, which fuses multi-view features hierarchically.
A comparison of the architectures of our deep fusion network and early/late fusion networks are shown in Fig.~\ref{fig:fusion}.
For a network that has $L$ layers, \emph{early fusion} combines features $\{f_v \}$ from multiple views in the input stage:
\begin{equation}
f_L = \bH_L(\bH_{L-1}(\cdots \bH_1(f_{BV}\oplus f_{FV} \oplus f_{RGB})))
\end{equation}
$\{\bH_l, l=1,\cdots,L\}$ are feature transformation functions and $\oplus$ is a join operation (e.g., concatenation, summation).
In contrast, \emph{late fusion} uses seperate subnetworks to learn feature transformation independently and combines their outputs in the prediction stage:
\begin{equation}
\begin{aligned}
f_L = & (\bH^{BV}_L(\cdots \bH^{BV}_1(f_{BV}))) \oplus \\
      & (\bH^{FV}_L(\cdots \bH^{FV}_1(f_{FV}))) \oplus \\
      & (\bH^{RGB}_L(\cdots \bH^{RGB}_1(f_{RGB})))
\end{aligned}
\end{equation}

To enable more interactions among features of the intermediate layers from different views, we design the following \emph{deep fusion} process:
\begin{equation}
\begin{aligned}
f_0 = & f_{BV} \oplus f_{FV} \oplus f_{RGB} \\
f_l = & \bH^{BV}_{l}(f_{l-1}) \oplus \bH^{FV}_{l}(f_{l-1}) \oplus \bH^{RGB}_{l}(f_{l-1}), \\
& \forall l=1, \cdots, L
\end{aligned}
\end{equation}
We use element-wise mean for the join operation for deep fusion since it is more flexible when combined with drop-path training~\cite{fractalnet}.

%=================== ground ================
\begin{table*}[t!]
\vspace{-1mm}
\begin{center}
\begin{small}
\addtolength{\tabcolsep}{-0pt}
\begin{tabular}{|c|c||c|c|c||c|c|c|}
\hline
\multirow{2}{*}{Method}& \multirow{2}{*}{Data}& \multicolumn{3}{|c||}{IoU=0.5} & \multicolumn{3}{|c|}{IoU=0.7} \\
\cline{3-8}
& & Easy & Moderate & Hard & Easy & Moderate & Hard \\
\hline
Mono3D~\cite{XiaozhiCVPR16} & Mono & 30.5 & 22.39 & 19.16 & 5.22 & 5.19 & 4.13 \\
3DOP~\cite{XiaozhiNIPS15} & Stereo & 55.04 & 41.25 & 34.55 & 12.63 & 9.49 & 7.59 \\
\hline\hline
VeloFCN~\cite{velofcn} & LIDAR & 79.68 & 63.82 & 62.80 & 40.14 & 32.08 & 30.47 \\
Ours (BV+FV) & LIDAR & 95.74 & 88.57 & 88.13 & 86.18 & 77.32 & 76.33 \\
%\hline\hline
Ours (BV+FV+RGB) & LIDAR+Mono & {\bf 96.34} & {\bf 89.39} & {\bf 88.67} & {\bf 86.55} & {\bf 78.10} & {\bf 76.67} \\
\hline
\end{tabular}
\end{small}
\vspace{-2.5mm}
\caption{{\bf 3D localization performance:} Average Precision (AP$_{\text{loc}}$) (in \%) of bird's eye view boxes on KITTI \emph{validation} set. For Mono3D and 3DOP, we use 3D box regression~\cite{XiaozhiPAMI17} instead of 2D box regression used in the original implementation.}
\label{tab:ap_loc_val}
\end{center}
\vspace{-5mm}
\end{table*}

%=================== 3D ================
\begin{table*}[t!]
\vspace{-1mm}
\begin{center}
\begin{small}
\addtolength{\tabcolsep}{-0pt}
\begin{tabular}{|c|c||c|c|c||c|c|c||c|c|c|}
\hline
\multirow{2}{*}{Method}& \multirow{2}{*}{Data}& \multicolumn{3}{|c||}{IoU=0.25} & \multicolumn{3}{|c||}{IoU=0.5} & \multicolumn{3}{|c|}{IoU=0.7} \\
\cline{3-11}
& & Easy & Moderate & Hard & Easy & Moderate & Hard & Easy & Moderate & Hard \\
\hline
Mono3D~\cite{XiaozhiCVPR16} & Mono & 62.94 & 48.2 & 42.68 & 25.19 & 18.2 & 15.52 & 2.53 & 2.31 & 2.31 \\
3DOP~\cite{XiaozhiNIPS15} & Stereo & 85.49 & 68.82 & 64.09 & 46.04 & 34.63 & 30.09 & 6.55 & 5.07 & 4.1 \\
\hline\hline
VeloFCN~\cite{velofcn} & LIDAR & 89.04 & 81.06 & 75.93 & 67.92 & 57.57 & 52.56 & 15.20 & 13.66 & 15.98 \\
Ours (BV+FV) & LIDAR & 96.03 & 88.85 & 88.39 & 95.19 & 87.65 & 80.11 & 71.19 & 56.60 & 55.30 \\
Ours (BV+FV+RGB) & LIDAR+Mono & {\bf 96.52} & {\bf 89.56} & {\bf 88.94} & {\bf 96.02} & {\bf 89.05} & {\bf 88.38} & {\bf 71.29} & {\bf 62.68} & {\bf 56.56} \\
\hline
\end{tabular}
\end{small}
\vspace{-2.5mm}
\caption{{\bf 3D detection performance:} Average Precision (AP$_{\text{3D}}$) (in \%) of 3D boxes on KITTI {\it validation} set. For Mono3D and 3DOP, we use 3D box regression~\cite{XiaozhiPAMI17} instead of 2D box regression used in the original implementation.}
\label{tab:ap_3d_val}
\end{center}
\vspace{-4mm}
\end{table*}

\vspace{-2mm}
\paragraph{Oriented 3D Box Regression}
Given the fusion features of the multi-view network, we regress to \emph{oriented} 3D boxes from 3D proposals.
In particular, the regression targets are the 8 corners of 3D boxes: $\bt=(\Delta x_0, \cdots, \Delta x_7, \Delta y_0, \cdots, \Delta y_7, \Delta z_0, \cdots, \Delta z_7)$.
They are encoded as the corner offsets normalized by the diagonal length of the proposal box.
Despite such a 24-D vector representation is redundant in representing an oriented 3D box, we found that this encoding approach works better than the centers and sizes encoding approach.
Note that our 3D box regression differs from ~\cite{deepslidingshape} which regresses to axis-aligned 3D boxes.
In our model, the object orientations can be computed from the predicted 3D box corners.
We use a multi-task loss to jointly predict object categories and oriented 3D boxes.
As in the proposal network, the category loss uses cross-entropy and the 3D box loss uses smooth $\ell_1$.
During training, the positive/negative ROIs are determined based on the IoU overlap of brid's eye view boxes.
A 3D proposal is considered to be positive if the bird's eye view IoU overlap is above 0.5,  and negative otherwise.
During inference, we apply NMS on the 3D boxes after 3D bounding box regression.
We project the 3D boxes to the bird's eye view to compute their IoU overlap.
We use IoU threshold of 0.05 to remove redundant boxes, which ensures objects can not occupy the same space in bird's eye view.

\vspace{-3mm}
\paragraph{Network Regularization}
We employ two approaches to regularize the region-based fusion network: \emph{drop-path} training~\cite{fractalnet} and \emph{auxiliary losses}.
For each iteration, we randomly choose to do global drop-path or local drop-path with a probability of 50\%.
If global drop-path is chosen, we select a single view from the three views with equal probability.
If local drop-path is chosen, paths input to each join node are randomly dropped with 50\% probability.
We ensure that for each join node at least one input path is kept.
To further strengthen the representation capability of each view, we add auxiliary paths and losses to the network.
As shown in Fig.~\ref{fig:fusion_net}, the auxiliary paths have the same number of layers with the main network.
Each layer in the auxiliary paths shares weights with the corresponding layer in the main network.
We use the same multi-task loss, i.e. classification loss plus 3D box regression loss, to back-propagate each auxiliary path.
We weight all the losses including auxiliary losses equally.
The auxiliary paths are removed during inference.

\subsection{Implementation}
\paragraph{Network Architecture.}
In our multi-view network, each view has the same architecture.
The base network is built on the 16-layer VGG net~\cite{verydeep} with the following modifications:

\begin{itemize}[topsep=0pt, itemsep=0pt, parsep=0pt]
\item Channels are reduced to half of the original network.
\item To handle extra-small objects, we use feature approximation to obtain high-resolution feature map. In particular, we insert a 2x bilinear upsampling layer before feeding the last convolution feature map to the 3D Proposal Network.
Similarly, we insert a 4x/4x/2x upsampling layer before the ROI pooling layer for the BV/FV/RGB branch.
\item We remove the 4th pooling operation in the original VGG network, thus the convolution parts of our network proceed 8x downsampling.
\item In the muti-view fusion network, we add an extra fully connected layer $fc8$ in addition to the original $fc6$ and $fc7$ layer.
\end{itemize}
We initialize the parameters by sampling weights from the VGG-16 network pretrained on ImageNet.
Despite our network has three branches, the number of parameters is about 75\% of the VGG-16 network.
The inference time of the network for one image is around 0.36s on a GeForce Titan X GPU.

\vspace{-3mm}
\paragraph{Input Representation.}
In the case of KITTI, which provides only annotations for objects in the front view (around $90^\circ$ field of view), we use point cloud in the range of [0, 70.4] $\times$ [-40, 40] meters.
We also remove points that are out of the image boundaries when projected to the image plane.
For bird's eye view, the discretization resolution is set to 0.1m, therefore the bird's eye view input has size of 704$\times$800.
Since KITTI uses a 64-beam Velodyne laser scanner, we can obtain a 64$\times$512 map for the front view points.
The RGB image is up-scaled so that the shortest size is 500.

\vspace{-3mm}
\paragraph{Training.}
The network is trained in an end-to-end fashion.
For each mini-batch we use 1 image and sample 128 ROIs, roughly keeping 25\% of the ROIs as positive.
We train the network using SGD with a learning rate of 0.001 for 100K iterations.
Then we reduce the learning rate to 0.0001 and train another 20K iterations.

%=================== ablation ================
\begin{table*}[t!]
\vspace{-0mm}
\begin{center}
\begin{small}
\addtolength{\tabcolsep}{-0pt}
\begin{tabular}{|c||c|c|c||c|c|c||c|c|c|}
\hline
\multirow{2}{*}{Data}& \multicolumn{3}{|c||}{AP$_{\text{3D}}$ (IoU=0.5)} & \multicolumn{3}{|c||}{AP$_{\text{loc}}$ (IoU=0.5)} & \multicolumn{3}{|c|}{AP$_{\text{2D}}$ (IoU=0.7)} \\
\cline{2-10}
& Easy & Moderate & Hard & Easy & Moderate & Hard & Easy & Moderate & Hard \\
\hline
Early Fusion & 93.92 & 87.60 & 87.23 & 94.31 & 88.15 & 87.61 & 87.29 & 85.76 & 78.77 \\
Late Fusion & 93.53 & 87.70 & 86.88 & 93.84 & 88.12 & 87.20 & 87.47 & 85.36 & 78.66 \\
Deep Fusion w/o aux. loss & 94.21 & 88.29 & 87.21 & 94.57 & 88.75 & 88.02 & 88.64 & 85.74 & 79.06 \\
Deep Fusion w/ aux. loss & {\bf 96.02} & {\bf 89.05} & {\bf 88.38} & {\bf 96.34} & {\bf 89.39} & {\bf 88.67} & {\bf 95.01} & {\bf 87.59} & {\bf 79.90} \\
\hline
\end{tabular}
\end{small}
\vspace{-2.5mm}
\caption{{\bf Comparison of different fusion approaches:} Peformance are evaluated on KITTI {\it validation} set.}
\label{tab:ablation-fusion}
\end{center}
\vspace{-5mm}
\end{table*}

\begin{table*}[t!]
\vspace{-0mm}
\begin{center}
\begin{small}
\addtolength{\tabcolsep}{-0pt}
\begin{tabular}{|c||c|c|c||c|c|c||c|c|c|}
\hline
\multirow{2}{*}{Data}& \multicolumn{3}{|c||}{AP$_{\text{3D}}$ (IoU=0.5)} & \multicolumn{3}{|c||}{AP$_{\text{loc}}$ (IoU=0.5)} & \multicolumn{3}{|c|}{AP$_{\text{2D}}$ (IoU=0.7)} \\
\cline{2-10}
& Easy & Moderate & Hard & Easy & Moderate & Hard & Easy & Moderate & Hard \\
\hline
FV & 67.6 & 56.30 & 49.98 & 74.02 & 62.18 & 57.61 & 75.61 &  61.60 & 54.29 \\
RGB & 73.68 & 68.86 & 61.94 & 77.30 & 71.68 & 64.58 & 83.80 & 76.45 & 73.42 \\
BV & 92.30 & 85.50 & 78.94 & 92.90 & 86.98 & 86.14 & 85.00 & 76.21 & 74.80 \\
\hline
FV+RGB & 77.41 & 71.63 & 64.30 & 82.57 & 75.19 & 66.96 & 86.34 & 77.47 & 74.59 \\
FV+BV & 95.19 & 87.65 & 80.11 & 95.74 & 88.57 & 88.13 & 88.41 & 78.97 & 78.16 \\
BV+RGB & {\bf 96.09} & 88.70 & 80.52 & {\bf 96.45} & 89.19 & 80.69 & 89.61 & {\bf 87.76} & 79.76 \\
\hline
BV+FV+RGB & 96.02 & {\bf 89.05} & {\bf 88.38} & 96.34 & {\bf 89.39} & {\bf 88.67} & {\bf 95.01} & 87.59 & {\bf 79.90} \\
\hline
\end{tabular}
\end{small}
\vspace{-2.5mm}
\caption{{\bf An ablation study of multi-view features}: Peformance are evaluated on KITTI {\it validation} set.}
\label{tab:ablation}
\end{center}
\vspace{-8mm}
\end{table*}

\section{Experiments}

We evaluate our MV3D network on the challenging KITTI object detection benchmark~\cite{kitti}.
The dataset provides 7,481 images for training and 7,518 images for testing.
As the test server only evaluates 2D detection, we follow~\cite{XiaozhiNIPS15} to split the training data into \emph{training} set and \emph{validation} set, each containing roughly half of the whole training data.
We conduct 3D box evaluation on the validation set.
We focus our experiments on the car category as KITTI provides enough car instances for our deep network based approach.
Following the KITTI setting, we do evaluation on three difficulty regimes: \emph{easy}, \emph{moderate} and \emph{hard}.

\vspace{-2mm}
\paragraph{Metrics.}
We evaluate 3D object proposals using \emph{3D box recall} as the metric.
Different from 2D box recall~\cite{Hosang2015PAMI}, we compute the IoU overlap of two \emph{cuboids}.
Note that the cuboids are not necessary to align with the axes, i.e., they could be oriented 3D boxes.
In our evaluation, we set the 3D IoU threshold to 0.25 and 0.5, respectively.
For the final 3D detection results, we use two metrics to measure the accuracy of 3D localization and 3D bounding box detection.
For 3D localization, we project the 3D boxes to the ground plane (i.e., bird's eye view) to obtain oriented bird's eye view boxes.
We compute Average Precision (AP$_{\text{loc}}$) for the bird's eye view boxes.
For 3D bounding box detection, we also use the Average Precision (AP$_{\text{3D}}$) metric to evaluate the full 3D bounding boxes.
Note that both the bird's eye view boxes and the 3D boxes are oriented, thus object orientations are implicitly considered in these two metrics.
We also evaluate the performance of 2D detection by projecting the 3D boxes to the image plane.
Average Preicision (AP$_{\text{2D}}$) is also used as the metric.
Following the KITTI convention, IoU threshold is set to 0.7 for 2D boxes.

\vspace{-3mm}
\paragraph{Baslines.}
As this work aims at 3D object detection, we mainly compare our approach to LIDAR-based methods VeloFCN~\cite{velofcn}, 3D FCN~\cite{li3dfcn}, Vote3Deep~\cite{vote3deep} and Vote3D~\cite{vote3d}, as well as image-based methods 3DOP~\cite{XiaozhiNIPS15} and Mono3D~\cite{XiaozhiCVPR16}.
For fair comparison, we focus on two variants of our approach, i.e., the purely LIDAR-based variant which uses bird's eye view and front view as input ({\bf BV+FV}), and the multimodal variant which combines LIDAR and RGB data ({\bf BV+FV+RGB}).
For 3D box evaluation, we compare with VeloFCN, 3DOP and Mono3D since they provide results on the validation set.
For 3D FCN, Vote3Deep and Vote3D, which have no results publicly available, we only do comparison on 2D detection on the test set.

%=============== 2D test ============
\begin{table}[t!]
%\vspace{-1mm}
\begin{center}
\begin{small}
\addtolength{\tabcolsep}{-3pt}
\begin{tabular}{|c|c||c|c|c|}
\hline
Method & Data & Easy & Mod. & Hard \\
\hline\hline
Faster R-CNN~\cite{fasterrcnn} & Mono & 87.90 & 79.11 & 70.19 \\
Mono3D~\cite{XiaozhiCVPR16} & Mono & 90.27 & 87.86 & 78.09 \\
3DOP~\cite{XiaozhiNIPS15} & Stereo & 90.09 & 88.34 & 78.79 \\
MS-CNN~\cite{mscnn} & Mono & 90.46 & 88.83 & 74.76 \\
SubCNN~\cite{subcnn} & Mono & 90.75 & 88.86 & 79.24 \\
SDP+RPN~\cite{sdp, fasterrcnn} & Mono & 89.90 & 89.42 & 78.54 \\
%DPM-C8B1~\cite{ref:Yebes14iv} & Stereo & 74.33 & 60.99 & 47.16  \\
\hline\hline
Vote3D~\cite{vote3d} & LIDAR & 56.66  & 48.05  & 42.64  \\
VeloFCN~\cite{velofcn} & LIDAR & 70.68 & 53.45 & 46.90 \\
Vote3Deep~\cite{vote3deep} & LIDAR & 76.95 & 68.39 & 63.22 \\
3D FCN~\cite{li3dfcn} & LIDAR & 85.54 & 75.83 & 68.30 \\
Ours (BV+FV) & LIDAR & 89.80 & 79.76 & 78.61 \\
Ours (BV+FV+RGB) & LIDAR+Mono & {\bf 90.37} & {\bf 88.90} & {\bf 79.81} \\
\hline
\end{tabular}
\end{small}
\vspace{-2.5mm}
\caption{{\bf 2D detection performance:} Average Precision (AP$_{\text{2D}}$) (in \%) for car category on KITTI \emph{test} set. Methods in the first group optimize 2D boxes directly while the second group optimize 3D boxes.}
\label{tab:ap_test}
\end{center}
\vspace{-5mm}
\end{table}

\begin{figure*}[t!]
\begin{center}
\vspace{-1mm}
\begin{tabular}{ccc}
\includegraphics[width=0.3\linewidth,trim = 0mm 0mm 0mm 65mm, clip]{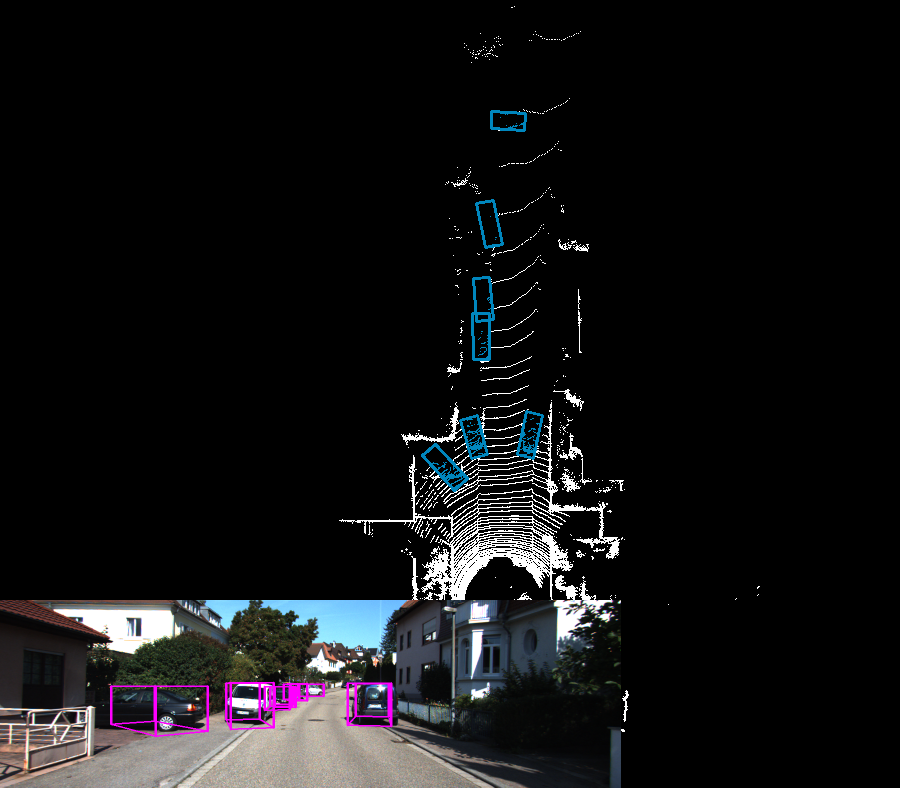}&
\includegraphics[width=0.3\linewidth,trim = 0mm 0mm 0mm 65mm, clip]{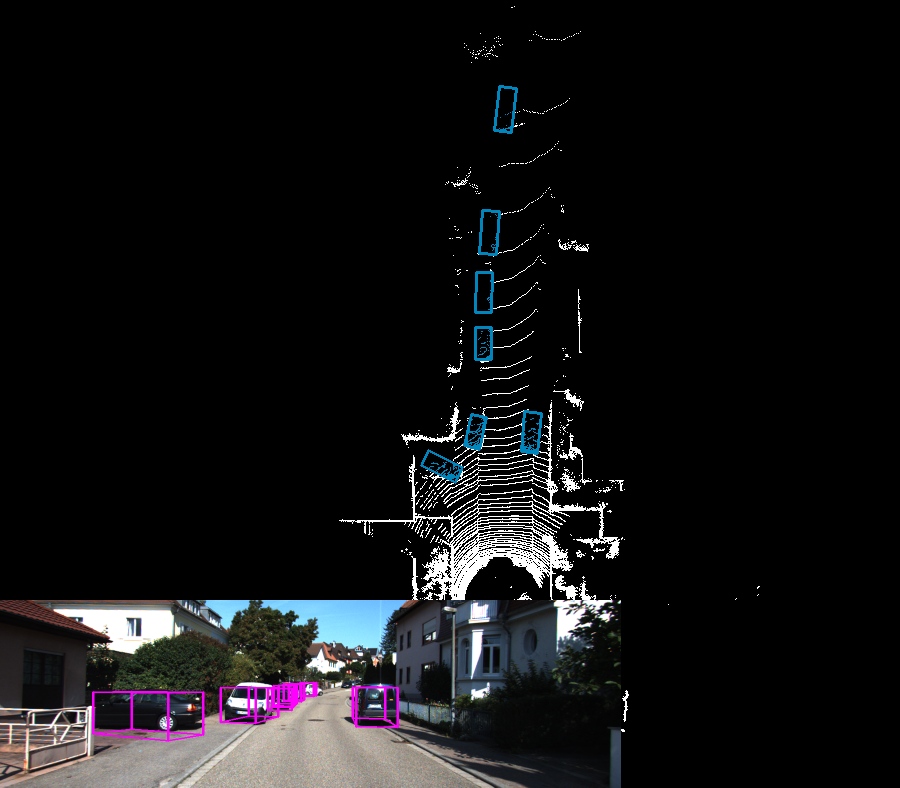}&
\includegraphics[width=0.3\linewidth,trim = 0mm 0mm 0mm 65mm, clip]{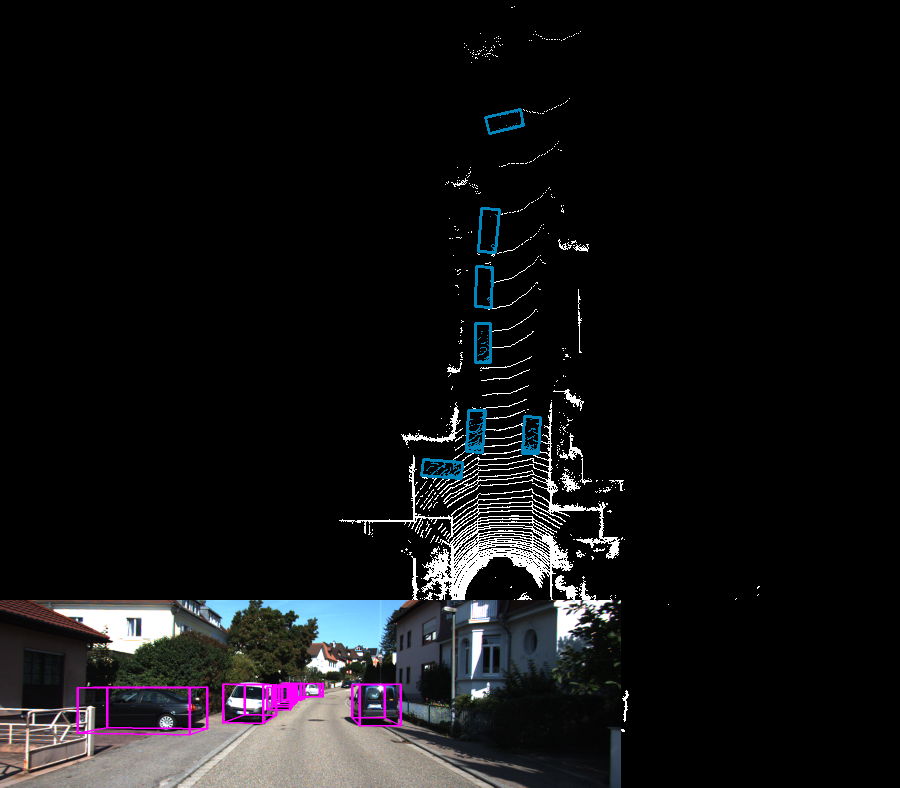}\\
\includegraphics[width=0.3\linewidth,trim = 0mm 0mm 0mm 70mm, clip]{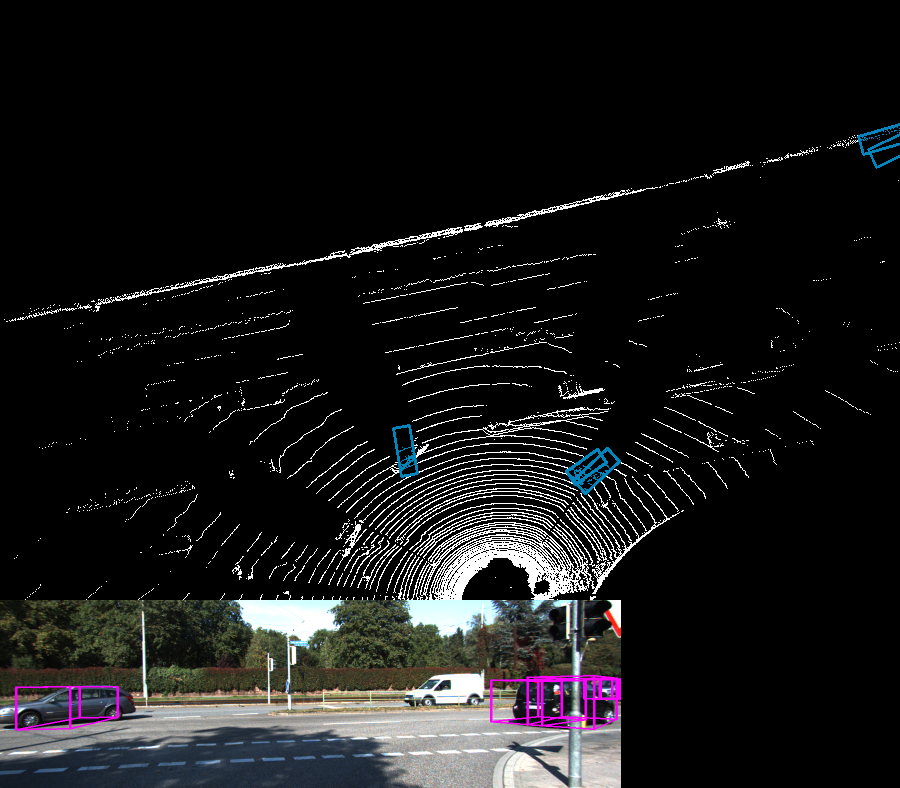}&
\includegraphics[width=0.3\linewidth,trim = 0mm 0mm 0mm 70mm, clip]{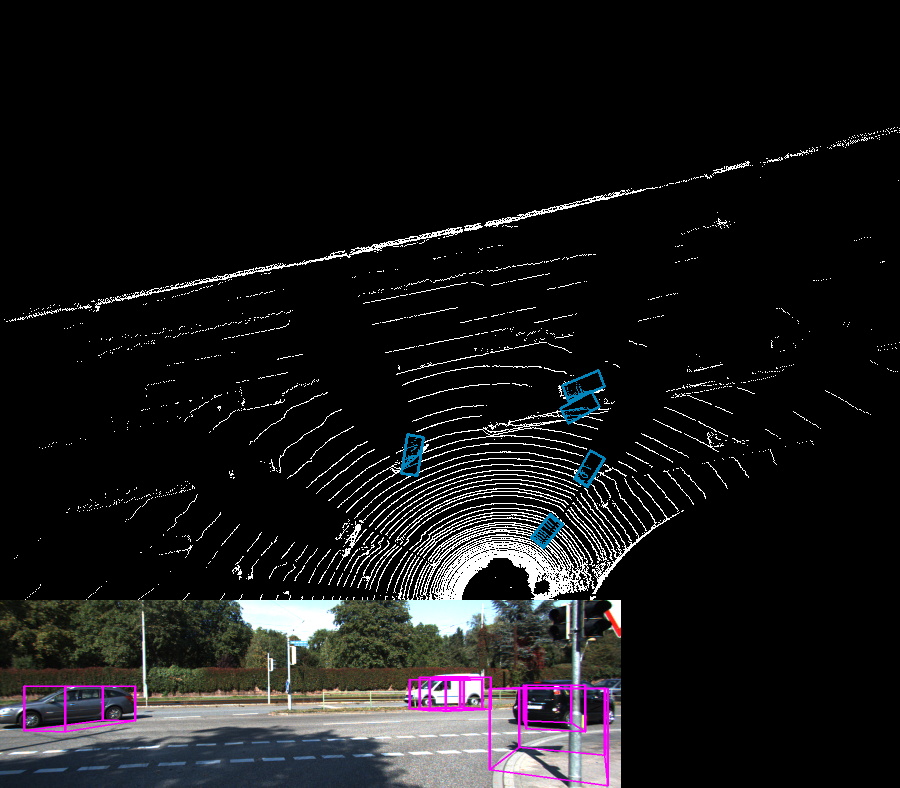}&
\includegraphics[width=0.3\linewidth,trim = 0mm 0mm 0mm 70mm, clip]{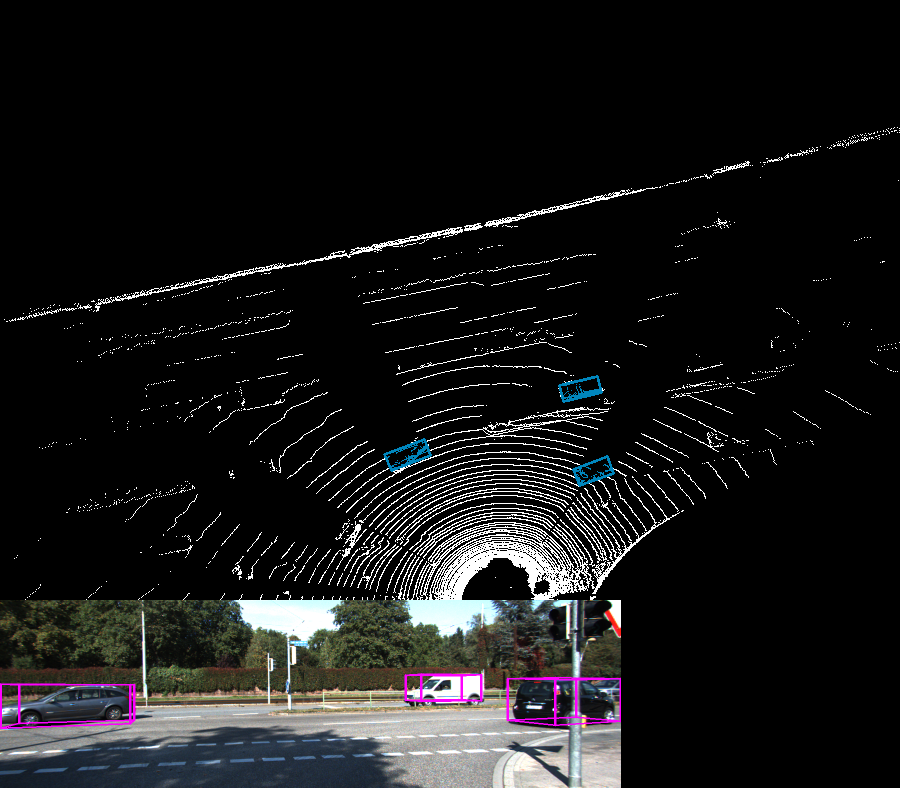}\\
3DOP~\cite{XiaozhiNIPS15} & VeloFCN~\cite{velofcn} & Ours \\
\end{tabular}
\vspace{-2.5mm}
\caption{\textbf{Qualitative comparisons of 3D detection results:} 3D Boxes are projected to the bird's eye view and the images.}
\label{fig:vis}
\end{center}
\vspace{-5mm}
\end{figure*}

\vspace{-4mm}
\paragraph{3D Proposal Recall.}
3D box recall are shown in Fig.~\ref{fig:3d-recall}.
We plot recall as a function of IoU threshold using 300 proposals.
Our approach significantly outperforms 3DOP~\cite{XiaozhiNIPS15} and Mono3D~\cite{XiaozhiCVPR16} across all the IoU thresholds.
Fig.~\ref{fig:3d-recall} also shows 3D recall as a function of the proposal numbers under IoU threshold of 0.25 and 0.5, respectively.
Using only 300 proposals, our approach obtains {\bf 99.1\%} recall at IoU threshold of 0.25 and {\bf 91\%} recall at IoU of 0.5.
In contrast, when using IoU of 0.5, the maximum recall that 3DOP can achieve is only 73.9\%.
The large margin suggests the advantage of our LIDAR-based approach over image-based methods.

\vspace{-4mm}
\paragraph{3D Localization.}
We use IoU threshold of 0.5 and 0.7 for 3D localization evaluation.
Table~\ref{tab:ap_loc_val} shows AP$_{\text{loc}}$ on KITTI \emph{validation} set.
As expected, all LIDAR-based approaches performs better than stereo-based method 3DOP~\cite{XiaozhiNIPS15} and monocular method Mono3D~\cite{XiaozhiCVPR16}.
Among LIDAR-based approaches, our method (BV+FV) outperforms VeloFCN~\cite{velofcn} by {\bf $\sim$25\%} AP$_{\text{loc}}$ under IoU threshold of 0.5.
When using IoU=0.7 as the criteria, our improvement is even larger, achieving {\bf $\sim$45\%} higher AP$_{\text{loc}}$ across easy, moderate and hard regimes.
By combining with RGB images, our approach is further improved.
We visualize the localization results of some examples in Fig.~\ref{fig:vis}.

\vspace{-4mm}
\paragraph{3D Object Detection.}
For the 3D overlap criteria, we focus on 3D IoU of 0.5 and 0.7 for LIDAR-based methods.
As these IoU thresholds are rather strict for image-based methods, we also use IoU of 0.25 for evaluation.
As shown in Table~\ref{tab:ap_3d_val}, our ``BV+FV" method obtains {\bf $\sim$30\%} higher AP$_{\text{3D}}$ over VeloFCN when using IoU of 0.5, achieving 87.65\% AP$_{\text{3D}}$ in the \emph{moderate} setting.
With criteria of IoU=0.7, our multimodal approach still achieves  71.29\% AP$_{\text{3D}}$ on \emph{easy} data.
In the \emph{moderate} setting, the best AP$_{\text{3D}}$ that can be achieved by 3DOP using IoU=0.25 is 68.82\%, while our approach achieves 89.05\% AP$_{\text{3D}}$ using IoU=0.5.
Some 3D detectioin results are visualized in Fig.~\ref{fig:vis}.

\vspace{-3mm}
\paragraph{Ablation Studies.}
We first compare our deep fusion network with early/late fusion approaches.
As commonly used in literature, the join operation is instantiated with concatenation in the early/late fusion schemes.
As shown in Table~\ref{tab:ablation-fusion}, early and late fusion approaches have very similar performance.
Without using auxiliary loss, the deep fusion method achieves $\sim$0.5\% improvement over early and late fusion approaches.
Adding auxiliary loss further improves deep fusion network by around 1\%.

To study the contributions of the features from different views, we experiment with different combination of the bird's eye view (BV), the front view (FV), and the RGB image (RGB).
The 3D proposal network is the same for all the variants.
Detailed comparisons are shown in Table~\ref{tab:ablation}.
If using only a single view as input, the bird's eye view feature performs the best while the front view feature the worst.
Combining any of the two views can always improve over individual views.
This justifies our assumption that features from different views are complementary.
The best overal performance can be achieved when fusing features of all three views.

\vspace{-4mm}
\paragraph{2D Object Detection.}
We finally evaluate 2D detection performance on KITTI \emph{test} set.
Results are shown in Table~\ref{tab:ap_test}.
Among the LIDAR-based methods, our ``BV+FV" approach outperforms the recently proposed 3D FCN~\cite{li3dfcn} method by {\bf 10.31\%} AP$_{\text{2D}}$ in the \emph{hard} setting.
In overall, image-based methods usually perform better than LIDAR-based methods in terms of 2D detection.
This is due to the fact that image-based methods directly optimize 2D boxes while LIDAR-based methods optimize 3D boxes.
Note that despite our method optimizes 3D boxes, it also obtains competitive results compared with the state-of-the-art 2D detection methods.

\vspace{-4mm}
\paragraph{Qualitative Results.}
As shown in Fig.~\ref{fig:vis}, our approach obtains much more accurate 3D locations, sizes and orientation of objects compared with stereo-based method 3DOP~\cite{XiaozhiNIPS15} and LIDAR-based method VeloFCN~\cite{velofcn}.

\vspace{-2mm}

\section{Conclusion}

We have proposed a multi-view sensory-fusion model for 3D object detection in the road scene.
Our model takes advantage of both LIDAR point cloud and images.
We align different modalities by generating 3D proposals and projecting them to multiple views for feature extraction.
A region-based fusion network is presented to deeply fuse multi-view information and do oriented 3D box regression.
Our approach significantly outperforms existing LIDAR-based and image-based methods on tasks of 3D localization and 3D detection on KITTI benchmark~\cite{kitti}.
Our 2D box results obtained from 3D detections also show competitive performance compared with the state-of-the-art 2D detection methods.

\paragraph{Acknowledgements.} The work was supported by National Key Basic Research Program of China (No. 2016YFB0100900) and NSFC 61171113.

{\small
\bibliographystyle{ieee}
\bibliography{cxz}
}

\end{document}